\pdfoutput=1

\documentclass[11pt]{article}

\usepackage[]{ACL2023}

\usepackage{times}
\usepackage{latexsym}

\usepackage[T1]{fontenc}

\usepackage[utf8]{inputenc}

\usepackage{microtype}

\usepackage{inconsolata}

\usepackage{amsfonts}
\usepackage{amsmath}
\usepackage{booktabs}
\usepackage{graphicx}
\usepackage[normalem]{ulem}
\useunder{\uline}{\ul}{}
\usepackage{caption}
\usepackage{subcaption}
\usepackage{CJKutf8}
\usepackage{multirow}
\usepackage{stfloats}
\usepackage{pdfpages}
\definecolor{mypink}{rgb}{0.858, 0.188, 0.478}

\title{Songs Across Borders: Singable and Controllable Neural Lyric Translation}

\author{
Longshen Ou \and Xichu Ma \and Min-Yen Kan \and Ye Wang \\
        National University of Singapore \\ 
    \texttt{\{longshen, ma\_xichu, kanmy, wangye\}@comp.nus.edu.sg}        
}

\begin{document}
\begin{CJK}{UTF8}{gkai}
\maketitle

\begin{abstract}

The development of general-domain neural machine translation (NMT) methods has advanced significantly in recent years, but the lack of naturalness and musical constraints in the outputs makes them unable to produce singable lyric translations. This paper bridges the singability quality gap by formalizing lyric translation into a constrained translation problem, converting theoretical guidance and practical techniques from translatology literature to prompt-driven NMT approaches, exploring better adaptation methods, and instantiating them to an English-Chinese lyric translation system. Our model achieves 99.85\%, 99.00\%, and 95.52\% on length accuracy, rhyme accuracy, and word boundary recall. In our subjective evaluation, our model shows a 75\% relative enhancement on overall quality, compared against naive fine-tuning\footnote{Code available at \\ \url{https://github.com/Sonata165/ControllableLyricTranslation}}. 

\end{abstract}
\section{Introduction}

\begin{figure*}[!b]

\includegraphics[width=1 \linewidth]{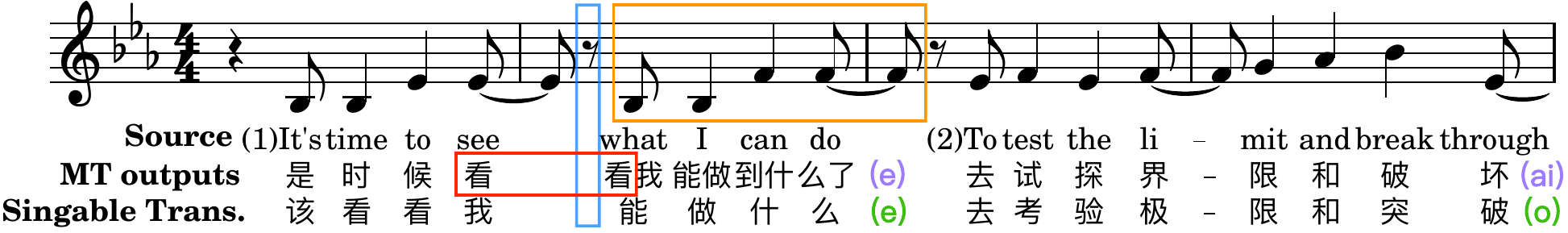}
\centering
\caption{Translation comparison of a general-domain NMT system (2nd row), already been adapted with parallel lyric data, versus a singable translation (3rd row).}
\label{fig:intro}

\end{figure*}

With the globalization of entertainment, it is becoming increasingly common for people to appreciate songs in foreign languages. Meanwhile, artists are internationalizing their work and building territories worldwide. Nevertheless, an unfriendly barrier exists between the artists and the audience: most commercial songs are not written in multiple languages. Worse still, most existing song translations entirely ignore the music constraints, rendering them unsingable alone with the music. As a result, the language barrier complicates the interaction between artists and their audience. 

Obtaining singable lyric translations can facilitate the globalization of the music publishing industry to further promote the growth of its \$5.9 billion USD market size \cite{market2022}. However, song translation is unusually difficult for human translators, due to music constraints and style requirements. If we can construct lyric-specific machine translation (MT) systems that can produce drafts that satisfy these constraints and requirements, the difficulty and cost of lyric translation will be largely reduced, as lyricists and translators can start with such automatic drafts and can focus on post-processing for quality and creativity. 

However, obtaining singable lyrics from MT systems is challenging. Figure~\ref{fig:intro} shows two sentences of lyrics from the song \textit{Let It Go}, together with an MT output and a singable translation. We observe a notable quality gap between them. While the MT output correctly translates the source, it ignores all the criteria that matter to make the output singable: 
(1) The second sentence of the MT outputs is unnatural because of incoherent vocabulary selection and lack of aesthetics. (2) Overcrowded syllables in the first sentence of the MT outputs force performers to break music notes in the orange box into multiple pieces to align them with lyrics. The rhythm pattern consequently diverges from the composer's intention. (3) The two-syllable word in the red box is situated across a musical pause (blue box), causing an unnatural pronunciation. (4) The end-syllables (purple text) are not of the same rhyme pattern, making the output miss a key chance for being poetic. 

In contrast, the singable translation in the third row outperforms the MT output in all four aspects, all while maintaining translation fidelity: it perfectly aligns with each musical note, has the same end-rhyme pattern for the two sentences (green text), a natural stop at the musical pause, and higher naturalness. These properties make it a significantly more performable translation.

To address these quality gaps to obtain singable lyric translations from neural machine translation (NMT) systems, we formalize singable lyric translation as an instance of constrained translation, identify useful constraints, and propose a language-pair independent approach that combines translatology theoretical guidance with prompt-driven NMT. Our contributions are:

\begin{itemize}
    \item We design an effective and flexible prompt-based solution for necessary word boundary position control that enhances the outputs' singability.
    \item We find that reverse-order decoding significantly contributes to the accuracy of prompt-based rhyme control. With this decoding strategy as the basis, we further design a rhyme ranking scheme to facilitate picking the best-suitable rhyme for translating input stanzas.
    \item  We conduct comparative studies of different prompt forms' effectiveness for controlling each aspect---length, rhyme, and necessary word boundary positions---and show the advantage of prompt-based control over control by modifying beam search.
    \item We show that adding back-translation of target-side monolingual data for fine-tuning is more effective in adapting the model to the lyric domain, compared with the more common practice of in-domain denoising pretraining.
\end{itemize}

\section{Related Work}

\paragraph{Lyric/Poetry Translation.} Designing domain-specific MT systems for poetic text translation, e.g., poetry and lyrics, is an emerging and underexplored topic in MT. Two previous works conducted pioneering research on lyrics \cite{guo2022automatic} and poetry \cite{ghazvininejad2018neural} translation separately by adopting a similar methodology of adjusting beam scores during beam search (referred to as \textit{biased decoding}) to encourage the generation of outputs with desired constraints. However, there is plenty of room for improvement. As will be shown in later sections, biased decoding not only fails at effectiveness of control, but also negatively impacts text quality and other simultaneously-controlled aspects. Additionally, the inclusion of controlling aspects is insufficiently comprehensive. For example, GagaST \cite{guo2022automatic} omits controls for rhyme, but rhyming is actually a critical desired property for song translations \cite{10.1093/ml/II.3.211}.

\paragraph{Lyric Generation.} 
Research on building lyric-specific language models shows the effectiveness of prompt-based control for outputs' length, rhyme, stress pattern, and theme \cite{li2020songnet, ma2021ai, xue2021deeprapper, ormazabal2022poelm, liu-etal-2022-chipsong}. However, several aspects remain to be enhanced. 

First, the prompts' forms vary: some works add prompts by additive embedding vectors \cite{li2020songnet, ma2021ai, xue2021deeprapper, liu-etal-2022-chipsong} and others by the prefix of input \cite{ormazabal2022poelm, liu-etal-2022-chipsong}. The lack of comparison makes it difficult to conclude the best prompt form for different control aspects. 

In addition, prior works did not control for some aspects in a well-designed manner. For example, \cite{liu-etal-2022-chipsong} enhances the music--lyric compatibility by controlling the number of syllables of \textit{each} word in the output. However, music constraints are usually not that tight so that such fine-level controlling might be unnecessary. Additionally, we found that unfitted rhyme prompts damage the output quality. However, we have not seen research suggesting how to choose the best suitable end-rhyme without naively traversing all possible rhyme prompts.

\paragraph{Translatology: Singable Translation of Songs.}
We attribute the inability of singable lyric translation from general-domain MT systems to the completely different goal of lyric translation compared with normal interlingual translation \cite{low2005pentathlon}: without considering the rhythm, note values, and stress patterns from music, song translations that seem good on paper may become awkward when singing. When the auditory perception is dominated by music \cite{golomb2005music}, the goal of translation is not again predominated by preserving the semantics of source text \cite{franzon2008choices}, but requires skilled handling of non-semantic aspects \cite{low2013songs} to attain the music--verbal unity, making it even an unusually complex task for human translators \cite{low2003singable}.

Theory and techniques from translatology provide valuable guidelines for our method design. Particularly, the ``Pentathlon Principle'' (\S\ref{sec:ctl_asp}) from \cite{low2003singable} is a widely accepted theoretical guidance to obtain singable song translations \cite{franzon2008choices, cheng2013singable, stopar2016mamma, si2017practical, opperman2018inter, sardina2021translation, pidhrushna2021functional}. In addition, some practical translation tricks have also been mentioned in \cite{low2003singable}, e.g., determining the last word first and from back to front when translating sentences in rhyme. 

\paragraph{Denoising Pretraining.}
The deficiency of in-domain data requires a powerful foundation model to ensure translation quality. We found large-scale denoising sequence-to-sequence pretraining \cite{lewis2019bart} a great candidate in our problem setting because it has been shown to be particularly effective in enhancing model's performance on text generation tasks such as summarization \cite{akiyama-etal-2021-hie} and translation \cite{liu2020multilingual, tang2020multilingual}, and also on domain-specific applications, e.g., \cite{yang2020generation, soper-etal-2021-bart, obonyo-etal-2022-exploring}. However, as indicated in \cite{liu2020multilingual}, the effectiveness of pretraining is related to the amount of monolingual data. In our case where in-domain data are relatively deficient, adopting the same strategy for adaptation might not be optimal.

\paragraph{Back-Translation.}
Back-translation (BT) and its variants can effectively boost the performance of NMT models \cite{sennrich2015improving, artetxe2017unsupervised, DBLP:conf/iclr/LampleCDR18}, and also show superior effectiveness in domain adaptation in low-resource settings \cite{hoang2018iterative, DBLP:conf/emnlp/WeiZCL20, zhang-etal-2022-iterative}. It is potentially a better adaptation method and may lead to higher output naturalness, which is required by singable translations.

\paragraph{Prompt-based Methods.}
Adding prompts during fine-tuning shows strong performance on lexical-constrained-MT \cite{susanto2020lexically, chousa-morishita-2021-input, wang-etal-2022-integrating}, as well as broad applicability on various controlling aspects such as output length \cite{lakew2019controlling} and the beginning word of output \cite{li2022prompt}. Compared to some earlier research that adds lexical constraints during beam search \cite{hokamp2017lexically, post2018fast}, 
the prompt based solution has a faster decoding speed and higher output quality \cite{susanto2020lexically}, hence might be the better option in our problem setting.

\section{Method}
\label{sec:method}

To bridge the gaps of previous research, we identify comprehensive controlling aspects from the translatology literature, propose prompt-based solutions for each aspect, and explore more effective foundation models and adaptation methods. 

\begin{table*}[t]

\centering
\resizebox{\textwidth}{!}{%
\begin{tabular}{@{}lll@{}}
\toprule
\multicolumn{1}{c}{\textbf{Aspects}} & \multicolumn{1}{c}{\textbf{Requirements}}                            & \multicolumn{1}{c}{\textbf{Our Actualization}}                                     \\ \midrule
(1) Singability                      & Outputs are suitable for singing with the given melodies.            & Enhance music-lyric compatibility by prompt-based necessary word boundary control. \\
(2) Rhythm                           & Outputs follow rhythm patterns in the music.                         & Prompt-based length (number of syllables) control.                                 \\
(3) Rhyme                            & Outputs fulfil certain rhyme patterns.                               & Prompt-based end-rhyme control and paragraph-level rhyme ranking.           \\
(4) Naturalness                      & Outputs read like lyrics originally composed in the target language. & Adapting with back-translation of in-domain target-side monolingual data.                        \\
(5) Sense                            & Outputs are fidelity to the meaning of source sentences.             & Large-scale general-domain pretraining.                                            \\ \bottomrule
\end{tabular}%
}
\caption{The ``pentathlon principle'' and the actualizations in our model.}
\label{tab:principle}
\end{table*}

\subsection{Controlling Aspects}
\label{sec:ctl_asp}

Are there some universal rules that we can adopt to obtain singable translations? We first rule out some prospective answers. Strictly keeping the positions of stressed syllables \cite{ghazvininejad2018neural} is inappropriate as stressing certain syllables is the property of stress-timed language. In contrast, syllable-timed languages, e.g., French and Mandarin, give syllables approximately equal prominence. Aligning the characters' tone with the melody \cite{guo2022automatic} is also not a good choice. On the one hand, this rule only applies to tonal languages. On the other hand, this rule is increasingly being ignored by the majority of songs composed in recent decades \cite{gao2017song}, indicating the marginalized importance of the intelligibility of songs, especially in pop\footnote{For example, according to Apple Music, 61 of the 2022 Top 100 Chinese pop songs are songs by Jay Chou, a Chinese artist famous for unintelligible songs.}.

To achieve a comprehensive and language-independent method, we define  ``singable translation'' as following the ``Pentathlon Principle'' from \cite{low2003singable}: that quality, singable translations are obtained by balancing five aspects---singability, rhythm, rhyme, naturalness, and sense. Table~\ref{tab:principle} lists these aspects and corresponding requirements, and how we actualize them in our model. Particularly, we identify (1)--(3) as the controlling aspects of our model and realize them with prompt-based control, while (4) and (5) are achieved from the perspectives of adaptation and pretraining.

\subsection{Problem Formulation}

We define the task that is tackled in this paper, {\it singable and controllable lyric translation}, as follows: given one line of lyrics $X$ in a source language $L_{src}$ and a set of desired properties of output sentence $\{l_{tgt}, r_{tgt}, \textbf{b}_{tgt}\}$, generating text translation $Y$ in target language $L_{tgt}$ for $X$ by modeling $P(Y|X,l_{tgt},r_{tgt}, \textbf{b}_{tgt})$, where (1) the total number of syllables of sentence $Y$ to be precisely equal to length constraint $l_{tgt}$; (2) $Y$ ends with a word that is in the same rhyme type of rhyme constraint $r_{tgt}$; 
(3) $Y$ has word boundaries---the positions between two consecutive syllables that belong to different words---in all locations indicated in necessary word boundary constraint $\textbf{b}_{tgt}$; (4) $Y$ is of maximal naturalness, and is fidelity to the sense of $X$.

\subsection{Prompt Methods for Controlling}
\label{sec:method_prompt}

Two types of special tokens are constructed as prompts for sentence-level control. For each sentence, the length and rhyme prompts are single token $\text{len}\_i$ and $\text{rhy}\_j$, indicating the desired number of syllables of the output is $i$ and that the desired end-rhyme type of output is $j$. The prompt for necessary word boundaries is a sequence of special tokens, $\textbf{bdr} = \{\text{bdr}\_0, \text{bdr}\_1\}^{\text{len}\_i}$, indicating the desired word boundary positions. 

During the training process, these prompts are derived from the analysis of target-side sentences, guiding the model towards generating sentences with corresponding properties. Consequently, there is no need for accompanying music during training. At the inference stage, prompts can be crafted from either music or source-side sentences. For an overview of the system workflow, please refer to Figures \ref{fig:workflow1} and \ref{fig:workflow2}.

We conducted a group of experiments to test three different prompt methods to determine the best one for each control aspect. They are (1) Enc-pref: prompts are injected into the encoder's input as a prefix. (2) Dec-pref: prompts are injected into the decoder's input as a prefix. (3) Dec-emb: prompts are embedded into a vector and added toward the decoder's input.

\subsection{Word Boundary Control}
\label{sec:bdr}
\begin{figure}[bt]
\centering
\begin{subfigure}{0.48\linewidth}
  \centering
  \includegraphics[width=\textwidth]{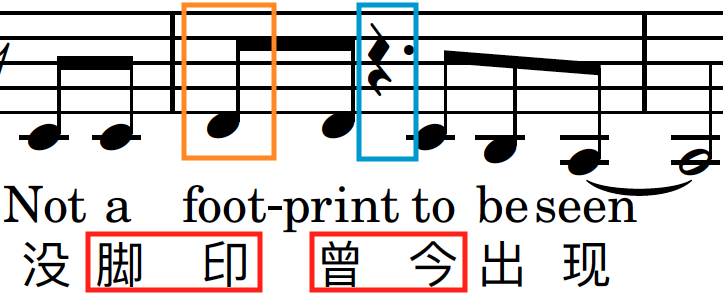}
  \vskip -2mm
  \caption{}
  \label{fig:bdr1}
\end{subfigure}
\hfill
\begin{subfigure}{0.48\linewidth}
  \centering
  \includegraphics[width=\textwidth]{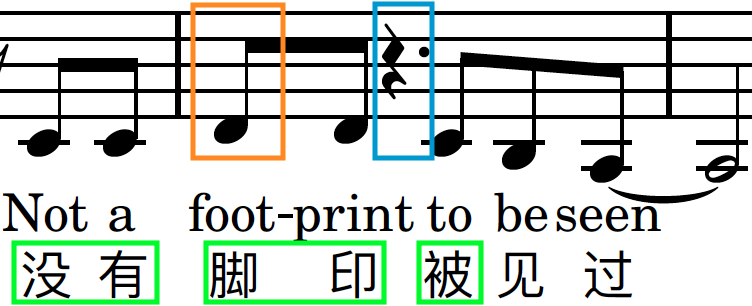}
  \vskip -2mm
  \caption{}
  \label{fig:bdr2}
\end{subfigure}
\vskip -0.1in
\caption{Demonstration of the necessity of word boundary control. Blue box: musical pauses; orange box: notes highlighted by downbeats; red box: words interrupted by musical pauses or highlighted notes; green box: words without interruption.}
\vskip -0.1in
\label{fig:bdr}
\end{figure}

Intra-word pause is a typical disfluency pattern of beginning language learners \cite{franco1998modeling}. However, improperly translated lyrics usually contain multi-syllable words that lies across musical pauses, as the blue box in Figure~\ref{fig:bdr}, so that the performer has to make awkward intra-word pauses while singing \cite{guo2022automatic}, causing a drop in pronunciation acceptability. Besides, we observe that positioning highlighted music notes, such as high notes or downbeats, as the orange box in Figure~\ref{fig:bdr}, onto a multi-syllable word's second or later syllables can bring similar adverse effects due to abrupt changes of pitch and tension\footnote{Stress-timed languages have another solution to this second problem, i.e., put a stressed syllable at the highlighted note. Here we discuss another generic solution.}. 

We address these issues by carefully designing the placement of \textit{word boundaries} in outputs, i.e., the points between two consecutive syllables that are from different words. Our aim is to ensure that word boundaries align precisely with the boundaries in music, i.e., the \textit{melody boundaries}, which occur \textit{at} musical pauses and \textit{before} highlighted notes (the blue and orange boxes in Figure~\ref{fig:bdr}). In this way, we achieve higher compatibility between the output sentences and the accompanying music, enhance the fluency and consistency of pronunciation during singing, and hence lead to the gain of singability.

This solution is achieved by prompt-based word boundary control. We use the prompt $\textbf{bdr}$ to represent melody boundary positions, indicating necessary word boundary positions. $\textbf{bdr}$ is a sequence of special tokens, and each token corresponds to one syllable in the output. There are two types of special interior tokens: \text{bdr}\_1 and \text{bdr}\_0, representing after the corresponding syllable ``there should be a word boundary'' and ``we do not care if there is a boundary'', respectively. At test time, $\textbf{bdr}$ is obtained from accompanying music and serves as additional inputs. A well-trained word-boundary-aware model can hence place word boundaries at the desired positions to achieve better music--lyric compatibility. 
For locations where $\text{bdr}\_0$ is present (``don't care''), the translation model operates unconstrained, maximizing translation naturalness. 

During training, length and rhyme prompts can be obtained directly from the target sentences in the training samples, but not again for necessary word boundary prompts because they have to be obtained from accompanying music which is absent in training. Nevertheless, we offer a solution: we randomly sample from all actual word boundary positions from the target-side text and use this sampled subset as ``pseudo ground truth'' to construct $\textbf{bdr}$ for training. 

\subsection{Reverse Order Decoding}

\subsubsection{Sentence-Level Control}

We imitate the process of human translators translating texts in rhyme: translating the last word first,
and from back to front, which is an old trick 
to keep rhyming patterns from being forced \cite{low2003singable}. We implement this by reverse-order decoding. During fine-tuning with parallel data, we reverse the word order of target-side text while retaining the source-side text unchanged. This approach minimally changes the structure and workflow of off-the-shelf translation models.

\subsubsection{Paragraph-Level Ranking}

Controllability alone is not enough. For a given input sentence, the rhyming usually only looks good in certain rhyme types but appears forced in others (see Appendix~\ref{sec:app_rhydif} for details). No matter how good the controllability is, the output quality will be severely damaged if an ill-fitting rhyme prompt is provided by the user. To avoid such problems, we need to determine the most suitable end-rhyme for translating one sentence, and further one paragraph consisting of multiple sentences. Previous research left this problem unsolved.

Fortunately, our reverse-order decoder simplifies the rhyme ranking process. During training, we use an additional special token $\text{rhy}\_0$ to nullify rhyme constraints for output. We achieve this by randomly converting a portion of each type of rhyme prompt to $\text{rhy}\_0$ during training. At inference time, for a given source sentence $X_i$ and prompts $l_{tgt}$, $r_{tgt}$ and $\textbf{b}_{tgt}$, we first use $\text{rhy}\_0$ as the rhyme prompt to do the first step of reverse-order decoding to obtain the end-word probability distribution,
\begin{align}
&P(y_{-1} | X, l_{tgt}, \textbf{b}_{tgt}, \text{rhy}\_0) \nonumber \\
&= [p(w_1), p(w_2), \ldots, p(w_v)],  
\label{eqn}
\end{align}
where the $v$ is the vocabulary size of the target language. Note that the $p(w_j)$ not only indicates the end-word probability, but also predicts output text quality and the likelihood of satisfaction of length and word boundary constraints of the rhyme-unconstrained model, from a greedy point of view. Intuitively, starting with tokens with low probabilities will pull down the corresponding beams' scores and degrade the output quality. On the contrary, sentences with higher quality can be obtained by starting decoding with $w_j$ with higher $p(w_j)$, and we achieve this by giving the model a rhyme prompt that guides it towards starting with such $w_j$. We sum up the probability in Eq.~\ref{eqn} within each rhyme type to obtain the rhyme distribution of given inputs,
\[
p_i = \sum_{Rhy(w_j) \in \text{rhyme }i} p(w_j)
\]
\begin{align*}
P(Rhy(Y) &| X, l_{tgt}, \textbf{b}_{tgt}, \text{rhy}\_0) \\
&= P(Rhy(y_{-1}) | X, l_{tgt}, \textbf{b}_{tgt}, \text{rhy}\_0) \\
&= [p_1, p_2, \ldots, p_{u}],
\end{align*}
where $Rhy(\cdot)$ is a map between a word or the end-word of a sentence to its rhyme type, $u$ is the number of rhyme types in the target language. For a certain rhyme type $i$, a higher $p_i$ value indicates a higher probability of successful rhyming and higher output quality. 

When translating a paragraph of lyrics, we have multiple sentences together with their corresponding length and boundary prompts as input:
\begin{align*}
&\mathbf{X}=[X_1, X_2, \ldots, X_n], \text{with prompts} \\
&[(l_{tgt_1}, \textbf{b}_{tgt_1}), (l_{tgt_2}, \textbf{b}_{tgt_2}), \ldots, (l_{tgt_n}, \textbf{b}_{tgt_n})].
\end{align*}
With the assumption that every sentence is of equal importance, we compute a normalized rhyme distribution for this paragraph by
\[
P(Rhy(Y_k)) = f(X_k, l_{tgt_k}, \textbf{b}_{tgt_k}, \text{rhy}\_0),
\]
\[
P(Rhy(\mathbf{Y})) = \text{softmax}(\sum_{k=1 }^{n}P(Rhy(Y_k)))
\]
where $f$ refers to the first step of reverse-order decoding. We then use $P(Rhy(\mathbf{Y}))$ as the rhyme ranking score of this paragraph to guide the rhyme selection.

\subsection{Utilizing Monolingual Data}

In-domain parallel data suffer from two issues. First, its amount is so limited that it is not comparable with general-domain data. Second, there are severe quality issues when target-side lyrics are translated by online communities, including wrong translation \cite{li2020problem}, creative treason \cite{zhang2022panni}, over-domestication \cite{xie2022guihua}, etc. 

To mitigate the issue of data quantity and quality, we seek help from target-side monolingual lyric data. Our approach involves incorporating back-translation \cite{sennrich2015improving} of target-side in-domain monolingual data to augment the parallel data for fine-tuning. 
To demonstrate its effectiveness, we conduct a comparative study with the adaptation method in \cite{guo2022automatic}, which performs sentence-level denoising pretraining \cite{lewis2019bart} with in-domain data after general-domain pretraining. 
\\

\begin{figure}[tb]
\centering
\begin{subfigure}{\linewidth}
  \centering
  \includegraphics[width=\textwidth]{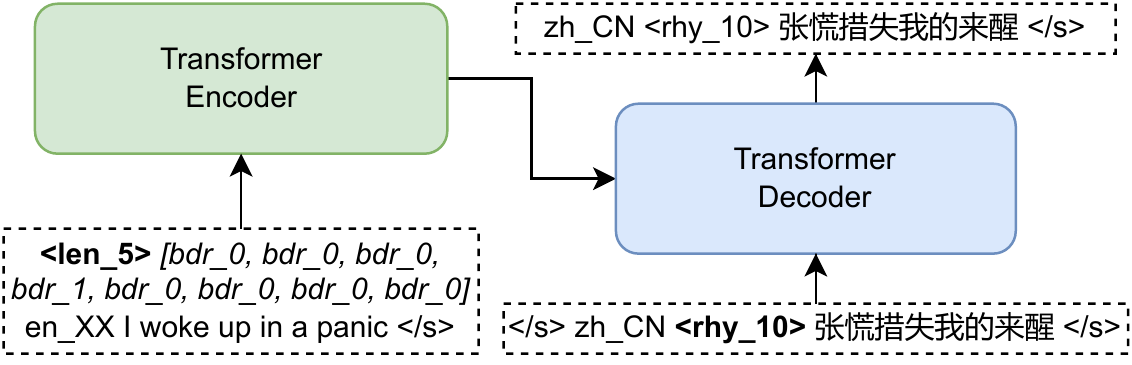}
  \vskip -2mm
  \caption{}
  \label{fig:system_sub}
\end{subfigure}

\begin{subfigure}{0.49\linewidth}
  \centering
  \includegraphics[width=\textwidth]{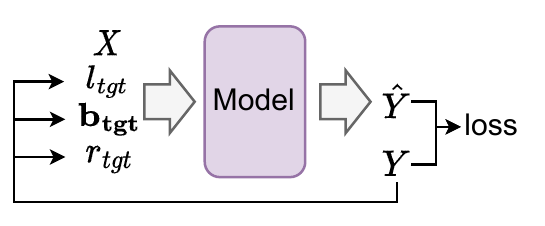}
  \vskip -2mm
  \caption{}
  \label{fig:workflow1}
\end{subfigure}
\hfill
\begin{subfigure}{0.49\linewidth}
  \centering
  \includegraphics[width=\textwidth]{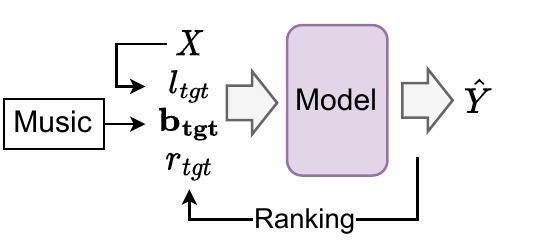}
  \vskip -2mm
  \caption{}
  \label{fig:workflow2}
\end{subfigure}%

\caption{(a): Structure of our English-to-Chinese lyric translation system. (b): Workflow of the fine-tuning stage. (c) Workflow of the inference stage. }
\label{fig:system}
\vskip -0.1in
\end{figure}

Taken together, these innovations form our final control method, which we can apply to any foundation model. In the evaluation that follows, we instantiate our techniques with Multilingual BART (refer to Figure~\ref{fig:system} for structure and workflow), producing the Singable Translation (Row~3) in Figure~\ref{fig:intro}. Additional case studies are featured in Appendix~\ref{app:case}.

\section{Experiment}
\label{sec:experiment}

We tested our methods with English--Chinese lyric translation. We obtained a small amount of parallel data (about 102K paired sentences after deduplication) by crawling data of both English--Chinese and Chinese--English pairs from an online lyric translation sharing 
platform\footnote{\url{https://lyricstranslate.com/}}. For target-side monolingual data, we adopted lyric data from three publicly-available datasets\footnote{\url{https://github.com/liuhuanyong/MusicLyricChatbot}}\footnote{\url{https://github.com/gaussic/Chinese-Lyric-Corpus}}\footnote{\url{https://github.com/dengxiuqi/ChineseLyrics}}, resulting in about 5.5M sentences after deduplication.
For details of dataset statistics and splits, data preprocessing, and back translation, please refers to Appendix~\ref{sec:app_data}.

\subsection{Model Configuration}
We adopted Multilingual BART \cite{liu2020multilingual} as the foundation model. We set the batch size to the largest possible value to fit into one NVIDIA A5000 GPU (24G), did simple searching for best learning rate, and kept the majority of other hyperparameters as default. For all experiments, models were first trained to converge on back-translated data, and fine-tuned with parallel data afterward. Please refer to Appendix~\ref{sec:app_mbart} for implementation details and hyperparameter setting.

\subsection{Evaluation}

The following metrics are used for objective evaluation: Sacre-BLEU \cite{post2018call}, TER \cite{snover2006study}, length accuracy (LA), rhyme accuracy (RA), and word boundary recall (BR). BLEU is a standard metric for various translation models. TER is also adopted because it directly reflects how much effort the lyricists need to spend to convert model outputs to perfectly singable lyrics. For length and rhyme control, we compare outputs' lengths and rhymes with desired constraints and compute the accuracy. For word boundary control, we first obtain outputs' word boundary locations using the Jieba tokenizer\footnote{\url{https://github.com/fxsjy/jieba}}, and then compute the recall value with the necessary word boundary prompts, indicating the ratio of satisfied desired word boundaries. 

For models that are constraint-aware for any controlling aspects, we conducted testing over two groups of experiments, as below:

{\bf Target as constraints (tgt-const):} For a given sentence pair, the length constraint is equal to the number of syllables of the target-side sentence; the rhyme constraint is equal to the rhyme category of the end-word of the target-side sentence; the boundary constraints are randomly sampled from word boundaries inside the target sentences. In this setting, the BLEU and TER scores represent the text quality directly.

{\bf Source as constraints (src-const):} For a given sentence pair, the length constraint is equal to the number of syllables of the source-side sentence; the rhyme constraint is randomly sampled from the real rhyme type distribution of lyrics in the target language, obtained from our monolingual dataset; the boundary constraints are randomly sampled from word boundaries inside the source sentences. This setting simulates real-world lyric translation cases and is more challenging.

\begin{table*}[t]
\centering
\begin{tabular}{@{}l|ccccc|cccc@{}}
\toprule
                                    & \multicolumn{5}{c|}{\textbf{Tgt-const}}                                            & \multicolumn{4}{c}{\textbf{Src-const}}                            \\
\multicolumn{1}{c|}{\textbf{Model}} & \textbf{BLEU↑} & \textbf{TER↓}  & \textbf{LA↑}   & \textbf{RA↑}   & \textbf{BR↑}   & \textbf{BLEU↑} & \textbf{LA↑}   & \textbf{RA↑}   & \textbf{BR↑}   \\ \midrule
Baseline                            & 21.71          & 70.04          & 20.54          & 37.49          & 62.28          & (21.71)              & 18.15          & 8.04           & 55.88          \\
Ours                                & \textbf{30.69} & \textbf{49.72} & \textbf{99.85} & \textbf{99.00} & \textbf{95.52} & 16.04          & \textbf{98.25} & \textbf{96.53} & \textbf{89.77} \\ \bottomrule
\end{tabular}
\caption{Results of our final model versus the baseline model. \textbf{Baseline}: mBART pretraining + finetuning with parallel data. \textbf{Ours}: mBART pretraining + finetuning with BT and parallel data + full constraints. LA, RA, BR refer to length accuracy, rhyme accuracy, and boundary recall, respectively. The best result is \textbf{bolded}. BLEU scores of baseline in the src-const setting, given in (parentheses), is not considered in the comparison in this and following tables.}
\label{tab:main}
\end{table*}

In src-const, we do not compare constrained models with unconstrained ones on BLEU or compute TER for outputs, as target-side sentences often possess distinct properties (e.g., \#~syllables) from prompts generated by source sentences, rendering them not the ground truth. Owing to the divergence between references and prompts, models with more constraints yield lower BLEUs, and TER in src-const fails to accurately reflect translation quality.

We compare our model with two baselines. The first is the unconstrained and un-adapted \textbf{Baseline} model presented in Table~2. The second is GagaST \cite{guo2022automatic}, which, to the best of our knowledge, is the only prior work on lyric translation. Due to data acquisition difficulties, we did not perform a model-level comparison with GagaST. Instead, we compared the effectiveness of their adaptation (in-domain denoising pre-training) and control method (biased decoding) with ours (BT and prompt-based control), and compare generation results through subjective evaluation.

\section{Results}
\label{sec:results}

Table~\ref{tab:main} shows the results of our final model. In the tgt-const setting, our model surpasses the baseline model on all objective aspects, not only with much higher BLEU and lower TER scores, but also achieves almost perfect length and rhyme accuracies and a competitive boundary recall score. The success of controlling length, rhyme, and word boundary while maintaining a high text quality enables our model to generate singable lyric translations. In addition, the controlling mechanism remains effective in the src-const setting, showing the generalizability of our methods.

\subsection{Unconstrained Models}

As in Table~\ref{tab:unconstrained}, both general-domain pretraining and in-domain fine-tuning are necessary to ensure translation quality. There are performance drops if any of the two components are canceled from the unconstrained model. Meanwhile, fine-tuning with back-translated in-domain monolingual data further contributes to the performance gain, showing higher adaptation effectiveness than in-domain pretraining. We also show BT's contribution to improving naturalness in \S 5.5. 

\begin{table}[tb]
\centering
\begin{tabular}{@{}lcc@{}}
\toprule
\textbf{Model}             & \textbf{BLEU↑} & \textbf{TER↓}  \\ \midrule
Transformer                & 8.97           & 84.92          \\
mBart w/o ft              & 16.44          & 84.64          \\
mBart pt + ft (baseline)   & 21.71          & 70.04          \\
+ In-domain denoise pt     & 22.18          & 68.61          \\
+ BT target side mono data & \textbf{25.53} & \textbf{64.22} \\ \bottomrule
\end{tabular}
\caption{Comparison of unconstrained models. Best result in \textbf{bold}.}
\label{tab:unconstrained}
\end{table}
\begin{table}[tb]
\centering
\resizebox{\columnwidth}{!}{%
\begin{tabular}{@{}l|ccc|cc@{}}
\toprule
\textbf{}                           & \multicolumn{3}{c|}{\textbf{Tgt-const}}             & \multicolumn{2}{c}{\textbf{Src-const}} \\
\multicolumn{1}{c|}{\textbf{Model}} & \textbf{BLEU↑} & \textbf{TER↓}  & \textbf{Len acc↑} & \textbf{BLEU↑}   & \textbf{Len acc↑}   \\ \midrule
Baseline                            & 21.32          & 69.89          & 20.78             & (21.32)          & 18.48               \\
Dec-emb                             & 22.06          & 67.11          & 24.18             & \textbf{21.42}   & 21.52               \\
Dec-pref                            & {\ul 22.16}    & {\ul 62.77}    & {\ul 82.94}       & 18.61            & {\ul 80.30}         \\
Enc-pref                            & \textbf{23.29} & \textbf{61.30} & \textbf{86.49}    & {\ul 19.12}      & \textbf{83.78}      \\ \bottomrule
\end{tabular}%
}
\caption{Comparison of prompt methods for length constraints. Decoding direction: normal. Best result in \textbf{bolded}, second best \underline{underlined}.}
\label{tab:len}
\end{table}
\begin{table}[t]
\centering
\resizebox{\columnwidth}{!}{%
\begin{tabular}{@{}l|cccc|ccc@{}}
\toprule
\textbf{}                           & \multicolumn{4}{c|}{\textbf{Tgt-const}}                           & \multicolumn{3}{c}{\textbf{Src-const}}           \\
\multicolumn{1}{c|}{\textbf{Model}} & \textbf{BLEU↑} & \textbf{TER↓}  & \textbf{LA↑}   & \textbf{RA↑}   & \textbf{BLEU↑} & \textbf{LA↑}   & \textbf{RA↑}   \\ \midrule
W/o ctrl                            & 21.48          & 62.65          & {\ul 86.87}    & 39.88          & (17.38)        & \textbf{84.61} & 8.19           \\
Dec-emb                             & 21.18          & 63.27          & 84.97          & 39.90          & \textbf{17.05} & 82.95          & 7.87           \\
Enc-pref                            & \textbf{23.30} & \textbf{58.57} & \textbf{87.06} & {\ul 85.77}    & {\ul 14.91}    & {\ul 83.97}    & {\ul 64.21}    \\
Dec-pref                            & {\ul 22.92}    & {\ul 58.84}    & 85.16          & \textbf{96.66} & 14.26          & 81.43          & \textbf{88.52} \\ \bottomrule
\end{tabular}%
}
\caption{Comparison of prompt methods for rhyme constraints, when controlling length and rhyme together with reverse-order decoding. The best result is marked in {\bf bold}, the second best \underline{underlined}.  W/o ctrl: length-control-only model. }
\label{tab:rhyme}
\end{table}
\begin{table}[tb]
\centering
\resizebox{\columnwidth}{!}{%
\begin{tabular}{@{}l|ccccc|cccc@{}}
\toprule
                                    & \multicolumn{5}{c|}{\textbf{Tgt-const}}                                            & \multicolumn{4}{c}{\textbf{Src-const}}                            \\
\multicolumn{1}{c|}{\textbf{Model}} & \textbf{BLEU↑}  & \textbf{TER↓}   & \textbf{LA↑}    & \textbf{RA↑}    & \textbf{BR↑}    & \textbf{BLEU↑}  & \textbf{LA↑}    & \textbf{RA↑}    & \textbf{BR↑}    \\ \midrule
W/o ctrl.                           & 29.60          & 51.02          & 99.40          & \textbf{99.20} & 75.20          & (16.57)        & 97.80          & {\ul 96.81}    & 58.49          \\
Dec-emb                             & \textbf{30.86} & {\ul 49.93}    & \textbf{99.85} & {\ul 99.15}    & {\ul 94.19}    & 15.84          & {\ul 97.99}    & 96.58          & {\ul 87.52}    \\
Dec-pref                            & 30.24          & 50.44          & 99.78          & 99.12          & 81.37          & \textbf{16.48} & 97.93          & \textbf{96.95} & 72.36          \\
Enc-pref                            & {\ul 30.73}    & \textbf{49.91} & {\ul 99.79}    & 98.93          & \textbf{94.96} & {\ul 15.88}    & \textbf{98.09} & 96.61          & \textbf{89.62} \\ \bottomrule
\end{tabular}%
}
\caption{Comparison of prompt methods for word boundary constraints. Decoding direction: reverse. The best result in {\bf bold}, the second best, \underline{underlined}.  W/o ctrl: model with only length and rhyme control.}
\label{tab:boundary}
\end{table}
\begin{table}[t!]
\centering
\resizebox{\columnwidth}{!}{%
\begin{tabular}{@{}l|cccc|ccc@{}}
\toprule
                                    & \multicolumn{4}{c|}{\textbf{Tgt-const}}                           & \multicolumn{3}{c}{\textbf{Src-const}}           \\
\multicolumn{1}{c|}{\textbf{Model}} & \textbf{BLEU↑}  & \textbf{TER↓}   & \textbf{LA↑}    & \textbf{BR↑}    & \textbf{BLEU↑}  & \textbf{LA↑}    & \textbf{BR↑}    \\ \midrule
Length-only                         & {\ul 26.86}    & {\ul 56.48}    & {\ul 99.43}    & 73.31          & (20.91)        & {\ul 97.70}    & 60.62          \\
+ Biased dec                           & 17.19          & 68.68          & 87.14          & {\ul 75.60}    & {\ul 13.85}    & 84.92          & {\ul 65.51}    \\
+ Prompt                            & \textbf{27.21} & \textbf{56.07} & \textbf{99.77} & \textbf{95.22} & \textbf{16.04} & \textbf{98.25} & \textbf{89.77} \\ \bottomrule
\end{tabular}%
}
\caption{Comparison of prompt and biased decoding for word boundary control. Best in {\bf bold}; second best, \underline{underlined}.}
\label{tab:biased_bdr}
\end{table}
\begin{table}[t]
\centering
\resizebox{\columnwidth}{!}{%
\begin{tabular}{@{}cl|cccc|ccc@{}}
\toprule
\multicolumn{1}{l}{}        & \textbf{}    & \multicolumn{4}{c|}{\textbf{Tgt-const}}                           & \multicolumn{3}{c}{\textbf{Src-const}}           \\
\multicolumn{2}{c|}{\textbf{Model}}        & \textbf{BLEU↑}  & \textbf{TER↓}   & \textbf{LA↑}    & \textbf{RA↑}    & \textbf{BLEU↑}  & \textbf{LA↑}    & \textbf{RA↑}    \\ \midrule
\multicolumn{1}{c|}{}       & Len only     & {\ul 26.86}    & {\ul 56.48}    & \textbf{99.43} & 40.04          & (20.91)        & \textbf{97.70} & 8.44           \\
\multicolumn{1}{c|}{L-to-R} & + Biased dec & 24.77          & 59.68          & {\ul 98.50}    & {\ul 83.18}    & {\ul 18.58}    & {\ul 96.38}    & {\ul 80.90}    \\
\multicolumn{1}{c|}{}       & Dec-pref     & \textbf{28.81} & \textbf{52.04} & 98.25          & \textbf{94.88} & \textbf{18.82} & 96.21          & \textbf{84.00} \\ \midrule
\multicolumn{1}{c|}{}       & Len only     & 26.04          & {\ul 57.09}    & {\ul 98.95}    & 43.36          & (20.63)        & 96.85          & 8.41           \\
\multicolumn{1}{c|}{R-to-L} & + Biased dec & {\ul 26.45}    & 57.82          & 98.83          & {\ul 86.99}    & {\ul 16.68}    & {\ul 96.90}    & {\ul 79.28}    \\
\multicolumn{1}{c|}{}       & Dec-pref     & \textbf{29.59} & \textbf{50.95} & \textbf{99.25} & \textbf{99.23} & \textbf{16.89} & \textbf{97.60} & \textbf{96.80} \\ \bottomrule
\end{tabular}%
}
\caption{Comparison of rhyme control performance of biased decoding and prompt method. L-to-R: decode in normal order; R-to-L: decode in reverse order. In each group, the best result is marked by boldface, the second best is marked by underline. }
\label{tab:reverse}
\end{table}

\subsection{Best Prompt Methods}
\label{sec:compare}

We select the most effective prompt method for different controlling aspects in our final model. Here are the effectiveness comparisons. 

{\bf Length Control.}
As shown in Table~\ref{tab:len}, the encoder-side prefix is the best prompt method for length control, with the highest length accuracy and higher translation quality than dec-pref.

{\bf Rhyme Control.}
As shown in Table~\ref{tab:rhyme}, the decoder-side prefix is the best method for rhyme control, with a significantly higher rhyme accuracy than the second-best method encoder-side prefix. 

{\bf Word Boundary Control.}\footnote{BT data are not added to length and rhyme control experiments to maximize the performance differences of different methods, but are added in word boundary control experiments because boundary awareness is much slower to learn.}
As shown in Table~\ref{tab:boundary}, enc-pref is the best for word boundary control with much higher effectiveness than dec-pref. It has comparable performance with dec-emb in tgt-const, but shows stronger controllability in the src-const setting, indicating better generalizability.

\subsection{Prompt-Based Word Boundary Control}
As in Table~\ref{tab:biased_bdr}, prompt-based control is much more successful than biased decoding in word boundary control, not only achieving high boundary recall (95.22\% and 89.77\%) but also slightly raising the length accuracy and text quality. On the contrary, biased decoding contributes limited power to word boundary control with the expense of significant drops in text quality and length control accuracy.

\subsection{Prompt-Based Reverse-Order Decoding}

{\bf Prompt vs. Biased Decoding.} As in Table~\ref{tab:reverse}, the prompt-based method again shows higher effectiveness in rhyme control, while the biased decoding again negatively impacts text quality. As in Appendix~\ref{app:dis_beam}, the prompt-based control enables the model to adjust the expression of the entire sentence according to the given desired rhyme, achieving higher consistency, but the biased decoding sometimes abruptly changes the end-word to fulfill the constraint without considering whether it is compatible with input sentence and target-side context.

{\bf Normal vs. Reverse.} Reverse-order decoding further raise the performance of prompt-based rhyme control, but conversely, only brings marginal improvement to biased-decoding-based control. A possible explanation is the inability of biased decoding to handle polyphones (see Appendix~\ref{app:dis_beam}). We observed multiple cases where \textit{one of} the pronunciation of the end-word in its output does satisfy the rhyme requirement, but \textit{is not} the pronunciation in that context. On the contrary, the prompt-based control is aware of the whole target-side sentence, and hence better controllability is achieved.

\subsection{Human Evaluation}
\label{sec:human}
We employ five students from a local university with music performance or lyric composing backgrounds. We let participants evaluate outputs on five-point scales and take the average as the final score. Evaluations are from four dimensions: (1) \textit{sense}, whether the translation output retains the meaning of the input sentence; (2) \textit{naturalness}, whether the translation output sounds like lyrics composed initially in the target language; (3) \textit{music--lyric compatibility}, the degree of outputs and music match with each other and the consequent singability gain; (4) \textit{Singable Translation Score (STS)}, the overall quality as singable translations, a single-value metric considering the satisfaction of all five perspectives in the Pentathlon Principle (\S \ref{sec:ctl_asp})\footnote{Translation outputs are available at \\ \url{https://www.oulongshen.xyz/lyric\_translation}}.

\begin{table}[tb]
\centering
\resizebox{\columnwidth}{!}{%
\begin{tabular}{@{}l|ccc|c@{}}
\toprule
\multicolumn{1}{c|}{\textbf{Model}} & \textbf{Sense} & \textbf{Naturalness} & \textbf{Compatibility} & \textbf{STS} \\ \midrule
Baseline                            & 4.02           & 3.80                 & 2.53                   & 2.04                     \\
GagaST                              & 3.84           & 3.72                 & 4.01                   & 2.97                     \\
Ours                            & 3.95           & 3.78                 & \textbf{4.42}          & \textbf{3.57}            \\
- bdr                               & 3.91           & 3.72                 & 4.21                   & 3.46                     \\
\hspace{5mm}- rhy                               & 4.15           & \textbf{4.03}        & 4.21                   & 3.24                     \\
\hspace{10mm}- len                               & \textbf{4.36}  & 3.96                 & 2.64                   & 2.31                     \\ \bottomrule
\end{tabular}%
}
\caption{Subjective evaluation results. \textit{bdr}: word boundary control; \textit{rhy}: rhyme control; \textit{len}: length control.}
\label{tab:subjective}
\end{table}

Table~\ref{tab:subjective} shows the subjective evaluation results of baseline, GagaST \cite{guo2022automatic}, our model, and some ablated variants. On the STS metric, which is the ultimate goal of singable lyric translation, our model significantly outperforms the baseline and GagaST by 75.0\% and 20.2\%, showing its ability to generate singable translations. Besides, our model performs especially well on music--lyric compatibility, by 74.7\% and 10.2\% higher scores than the baseline and GagaST. In contrast, the baseline model performs worst on the two metrics. 

In addition, we show the contributions of different components by the ablated studies. The word boundary control raises music--lyric compatibility (+0.21) and overall quality (+0.11). The contribution from rhyme control is majorly on the overall quality part (+0.22), but with the expense of sense (-0.24) and naturalness (-0.31). Length control is the foundation of music--lyric compatibility (+1.57) and STS (+0.93), but with some expense of sense (-0.21). Adaptation with BT increases sense (+0.34) and naturalness (+0.16).


\section{Conclusion}
\label{sec:bibtex}

We discussed how to obtain singable translations with prompt-driven NMT systems with the guidance of translatology theories. Specifically, we used back-translation to enhance translation quality and naturalness. We compared the effectiveness of different prompt methods in different controlling aspects and showed their advantage over biased decoding. We designed an effective word boundary control approach and presented a training strategy without the help of music data. We demonstrated the effectiveness of reverse-order decoding in NMT models for rhyme control and showed how it helps users to choose the best suitable rhymes for a paragraph of source text.

This work does not explore more detailed prompt manipulation, such as using varied prompts for the same constraint or examining prompt order's impact on performance. We leave these investigations for future research.

\section*{Limitations}

The current system may require the user to have some music knowledge to compose the word boundary prompt from music. Hence, more efforts need to be made to fulfill this gap before such a system can operate fully automatically without the human user providing word boundary prompt themselves. 

We use the back-translation of mono-lingual data to augment the parallel training data, but the quality, especially the text style of back-translations has room to improve. Although we have tried using iterative BT to gradually refine the backward direction MT model to adapt its outputs to lyric style, we found some errors gradually accumulated in the back-translated data, which finally made our model perform unsatisfactorily for negative sentences, together with the decrease of controlling effectiveness. Further exploration is needed in this aspect.

Similar to chat text, lyrics are usually composed in short sentences. Sometimes it would be challenging to guarantee the consistency of style and meaning for different sentences, if the current sentence-level translation system are adopted. Hence, for building future lyric translation systems, it would be a better option to translate the lyrics directly at the paragraph level or document level.

\section*{Ethics Statement}
Our system will help facilitate the creation/re-creation of lyrics for song composers. In addition, although our system is implemented in the direction of English-to-Chinese, the controlling aspects and approaches are universal because we did not take any language-specific aspects into account; hence can be easily implemented in other language pairs. Besides, the method and system discussed in this paper are suitable for creating/re-creating singable song lyrics in languages beyond the original version. They also have the potential to benefit language learning by translating domestic languages into other languages the learner is studying and facilitating learning by singing. 

This methodology has limitations by putting the singability into priority. Translations from this system may sometimes not convey the exact meaning of the lyrics in the source language, causing misunderstanding in this case. For cases where conveying the original meaning is crucial, e.g., advertising and serious art songs, the translation outputs need to be checked and revised when necessary by the user before further usage.

For the training and evaluation of our system, all data is publicly available online. Specifically, Chinese Lyric Corpus\footnote{\url{https://github.com/gaussic/Chinese-Lyric-Corpus}} is a public GitHub repository with an MIT license. Lyricstranslate.com is a lyric translation sharing platform, where all parallel lyrics we obtained are publicly available in this website. We adhere to the rules specified in the website's robots.txt file when crawling. For all existing scientific artifacts used in this research, including datasets, models, and code, we ensure they are used in their original intended usage. For human evaluation, we collect evaluation scores without personal identifiers for subjective evaluation to ensure a fair comparison. We ensure that the questionnaire does not contain any offensive content. Please refer to Appendix~\ref{sec:sub_detail} for more details of subjective evaluation.

\section*{Acknowledgements}
This project was funded by research grant A-0008150-00-00 from the Ministry of Education, Singapore.

\bibliography{anthology,main}
\bibliographystyle{acl_natbib}

\appendix

\begin{table}[t!]
\centering
\resizebox{\columnwidth}{!}{%
\begin{tabular}{@{}cl|rrrr@{}}
\toprule
                                          &            & \textbf{Train} & \textbf{Validation} & \textbf{Test} & \textbf{Total} \\ \midrule
\multirow{2}{*}{\textbf{Back-translated}} & \#songs     & 142,796        & 104                 & 104           & 143,004        \\
                                          & \#sentences & 2,720,603      & 2,164               & 2,175         & 2,724,942      \\
\multirow{2}{*}{\textbf{Parallel}}        & \#songs     & 5,341          & 196                 & 201           & 5,738          \\
                                          & \#sentences & 102,177        & 4,011               & 4,006         & 110,194        \\ \bottomrule
\end{tabular}%
}
\caption{Dataset size of different splits.}
\label{tab:data}
\end{table}
\begin{figure*}[b!]
\includegraphics[width=0.9 \linewidth]{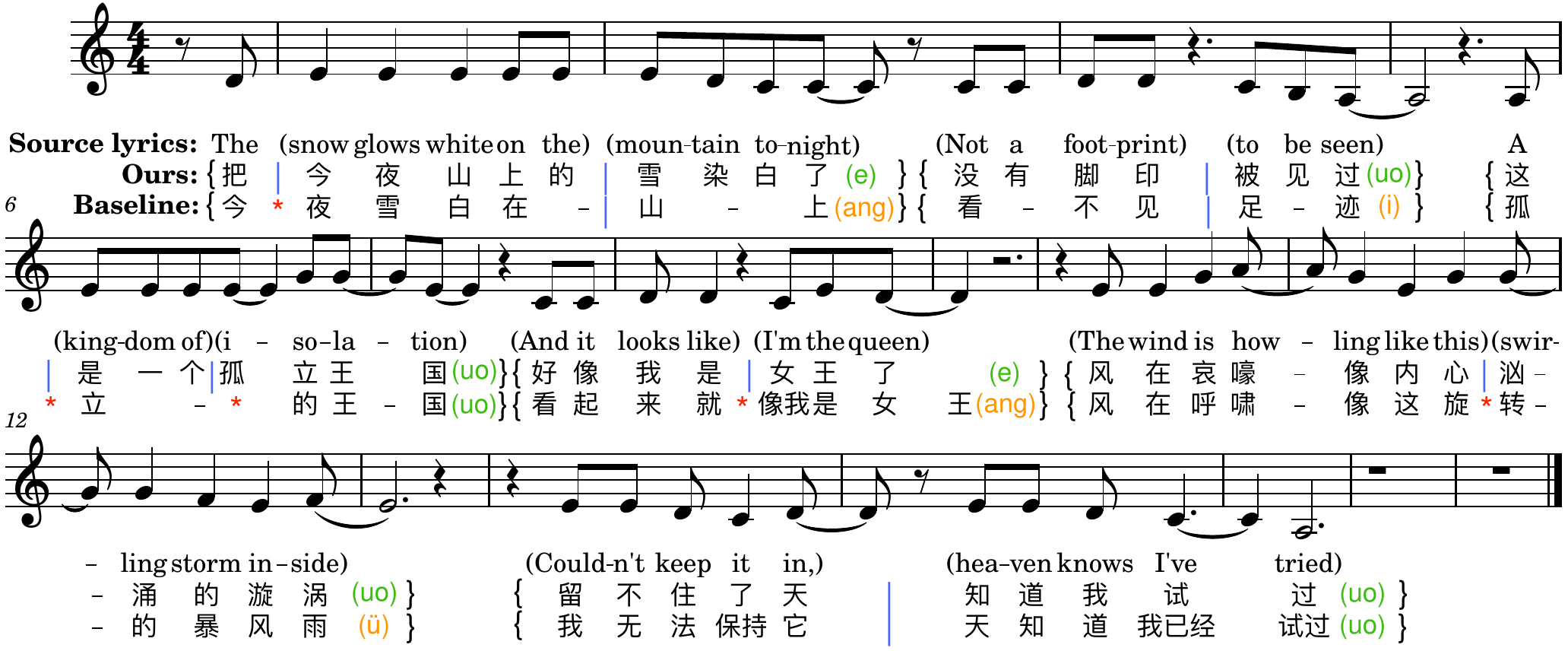}
\centering
\caption{Translation comparison of the our model and the baseline (mBART + finetuning with parallel data). Source lyrics are from the first verse of the song \textit{Let It Go}. Prompt: length equals to number of syllables of source text; 1st-ranked rhyme (type 2 \{o, e, uo\}); word boundary is extracted from melody, as shown in the source lyrics by parentheses. Sentence boundaries are marked by ``\{'' and ``\}''. Satisfied and unsatisfied rhymes are marked by {\color{green} green} and {\color{orange} orange} texts respectively. Satisfied and unsatisfied word boundaries are marked by {\color{blue}|} and {\color{red}*} respectively}
\label{fig:letitgo}
\end{figure*}
\begin{figure}[t]
\centering
\begin{subfigure}{0.9\linewidth}
    \centering
    \includegraphics[width=\textwidth]{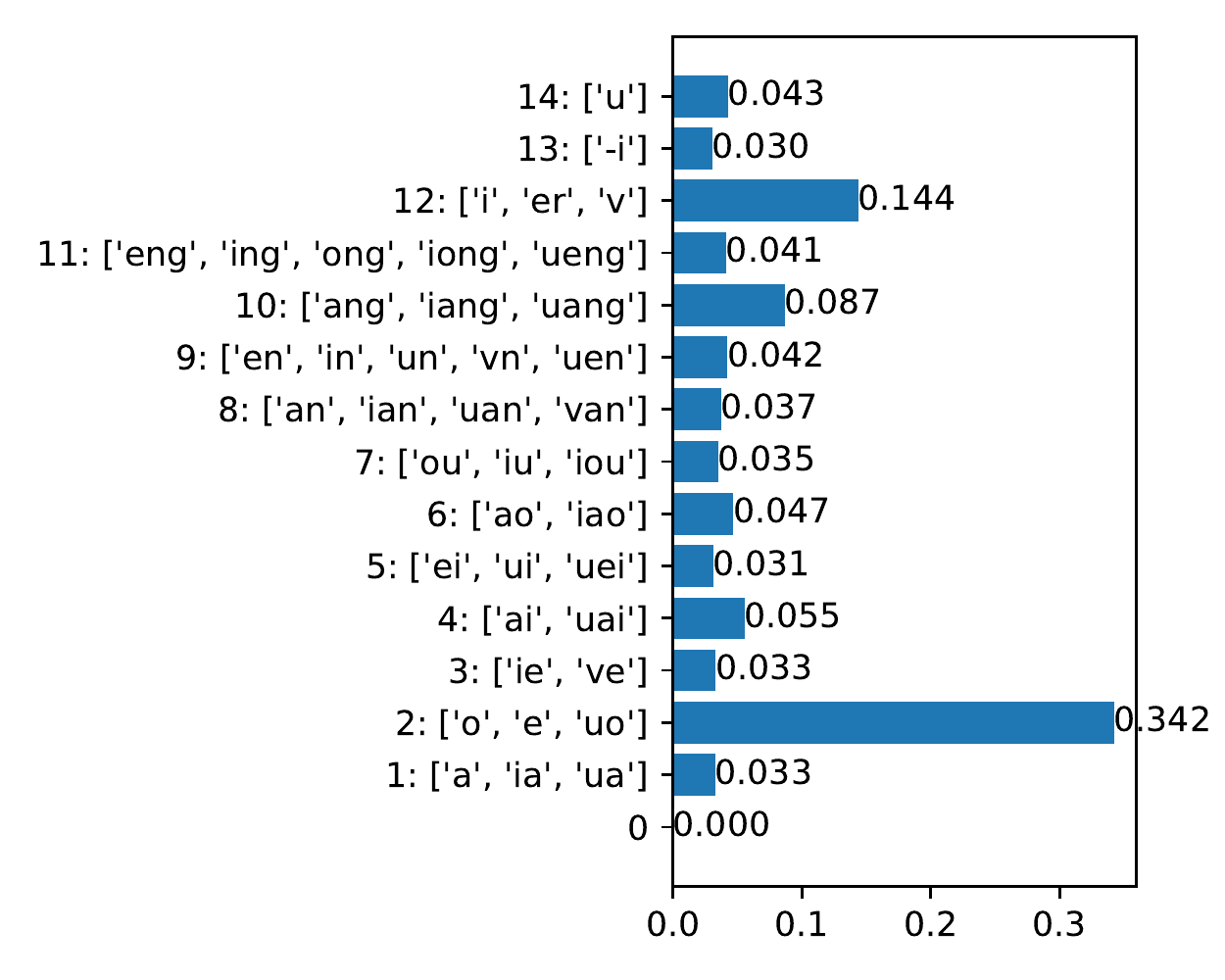}
    \vskip -4mm
    \caption{}
    \label{fig:rhy_dist}
\end{subfigure}
\vskip 2mm
\begin{subfigure}{0.44\linewidth}
    \centering
    \includegraphics[width=\textwidth]{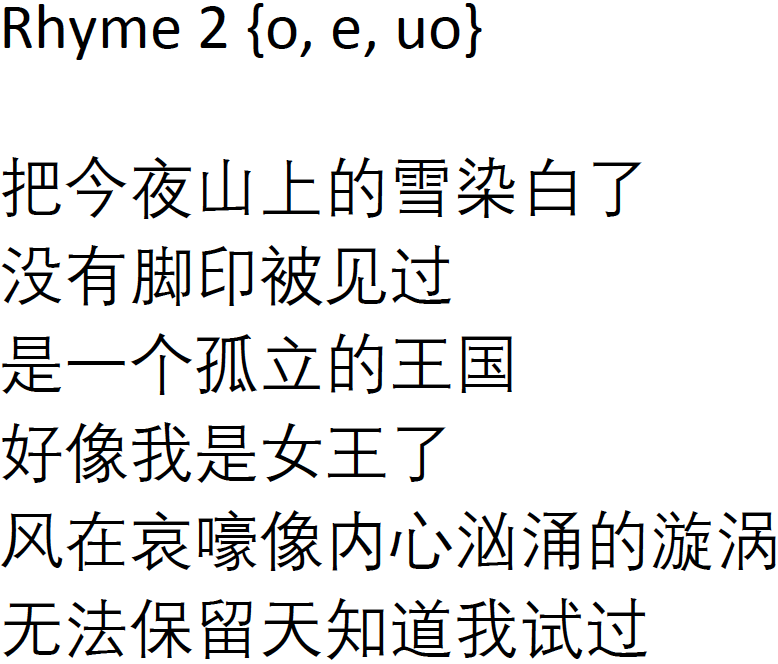}
    \vskip -2mm
    \caption{}
    \label{fig:rhy_dif1}
\end{subfigure}
\begin{subfigure}{0.44\linewidth}
    \centering
    \includegraphics[width=\textwidth]{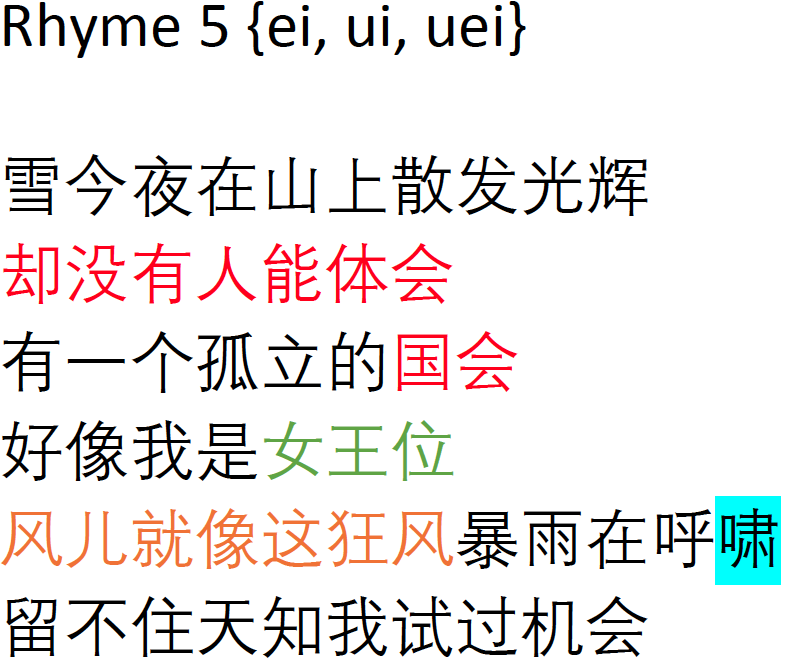}
    \vskip -2mm
    \caption{}
    \label{fig:rhy_dif2}
\end{subfigure}
\caption{(a): Rhyme ranking scores of different rhymes, when translating the the paragraph in Figure \ref{fig:letitgo}. (b) and (c): different translation output with different rhymes, using the rhyme with highest ranking score and with second lowest ranking score, respectively. Translation errors are marked in the right paragraph: wrong translation are marked with red, text marked in green does not conforms to target language grammar, orange text is repeated translation, highlighted word is in wrong rhyme.}
\label{fig:rhy_dif}
\vskip -0.1in
\end{figure}

\section{Data Preprocessing}
\label{sec:app_data}

\subsection{Dataset Details}
The monolingual lyric corpus from three sources includes lyrics data in Chinese, and vast majority of them are in pop genre. Lyrics of one song contains multiple lines. Each line usually corresponds to one utterance in singing. The length of each line is usually short. There are 8.6 Chinese characters each line on average. Only a few cases contains lines longer than 20 Chinese characters.

The crawled parallel lyrics contains two parts. For the first part, the lyrics are created in English originally, and translated to Chinese by online communities. The second part is composed in Chinese originally and translated to English. Similarly, most of them are in pop genre.

\subsection{Dataset Splitting}

Train/validation/test splitting is performed separately for BT and parallel data. Table~\ref{tab:data} shows the detailed statistics.

\subsection{Data Preprocessing}
We perform text normalization for all Chinese lyric text: all special symbols are removed; traditional characters are substituted with simplified characters\footnote{Follow the implementation of \url{https://github.com/liuhuanyong/MusicLyricChatbot/blob/master/process_data/langconv.py}}; sentences that are longer than 20 characters are removed; any duplicated sentences are removed. Finally, we split the datasets into train, validation, and test splits while ensuring no same songs exist in different splits.

For in-domain denoising pretraining experiments, text corrupting is performed by sentence-level mask prediction. There is one mask for each sentence. For the span of masks, for sentences with length in $(1, 3]$ and larger than 3, the mask span is sampled from a Poisson distribution with lambda equals 1 and 3, respectively.

\subsection{Back Translation}
For back translation, we adopt a Transformer trained with generic-domain Chinese-to-English data\footnote{\url{https://huggingface.co/Helsinki-NLP/opus-mt-zh-en}} to obtain sentence-level back translation.

\section{Implementation details}
\label{sec:app_mbart}

\paragraph{Model Configuration}
At the early stage of our experiment, we found that fine-tuning with generic-domain data does not help with the translation quality of lyrics. Hence we adopt mBART without general-domain fine-tuning as the starting point of training. For the unadapted general-domain model, we use mbart-large-50-one-to-many\footnote{\url{https://huggingface.co/facebook/mbart-large-50-one-to-many-mmt}}.

Our final model is obtained by fine-tuning mbart-large-50\footnote{\url{https://huggingface.co/facebook/mbart-large-50}} (\#param: 610,879,488) with both back-translated monolingual data and parallel data. The tokenizer is modified to be character-level on the Chinese side for better controlling effectiveness. The model is trained on one Nvidia A5000 GPU (24GB) for 10 epochs and 3 epochs on back-translation and parallel data, respectively, taking about 16 hours and 3 hours. The learning rate is set to 3e-5 and 1e-5, respectively, on BT and parallel data. They are the best value in \{1e-5, 3e-5, 1e-4\} for the baseline model on the two stages of training. Warm-up steps are set to 2500 and 300 for training with the BT and the parallel data. Dropout and label smoothing are set to 0. For decoding, beam-search with beam size 5 is adopted. The maximum output length is set to 30. All other hyperparameters remain as default values.

For the dec-emb experiments, instead of using sinusoidal encoding for prompts, we use learnable embedding to keep aligned with the positional embedding of mBART.

\paragraph{Length Prompt.}
We construct 20 length tokens for length control, $\text{len}\_{1}$ to $\text{len}\_{20}$ for translation output. According to the authors' observation, only an extremely tiny amount of lyrics in Mandarin have more than 20 characters in one line. 

\paragraph{Rhyme Prompt.}
\label{app:rhy}

For rhyme control, we adopt the Chinese 14-rhyme scheme\footnote{\url{https://github.com/korokes/chipsong}} for possible rhyme type, implemented as $\text{rhy}\_1$ to $\text{rhy}\_{14}$. There is a special token rhy\_0 representing ``no rhyme control''. This is achieved by randomly setting $1/15$ of each type of rhyme prompt to rhy\_0 during training.

\paragraph{Word Boundary Prompt}
We first sample a number $n$ from a categorical distribution with the ratio of 1:4:3:1 for 1, 2, 3, and 4 boundaries, and use $n' = min(\text{number of words}, n)$ as the number of bdr\_1 tokens. Then, we uniformly sample $n'$ times from all syllable boundary locations, without replacement, as the locations of these bdr\_1. After that, we initialize the prompt sequence as a sequence of bdr\_0 where the length of the sequence equals the number of syllables in the reference sentence. Finally, we substitute bdr\_0 with bdr\_1 for the sampled locations.

\section{More Case Studies}
\label{app:case}

\subsection{Model Outputs}

We show the translation comparison of the proposed model and the baseline model in Figure~\ref{fig:letitgo}. The outputs are perfect in the number of syllables and rhyme constraints. With the guidance of word boundary constraints, the output has much higher music-lyric compatibility than the baseline's output. For example, there is a downbeat lying on the note of the second word in the source lyrics, "snow", creating a melody boundary between the first and the second note. To get rid of pronunciation interruption, our system successfully places a word boundary here, avoiding the scenario where the second syllable of the word "今夜" is highlighted. Similarly, in the fourth sentence, our system places a word boundary at the place between the translation of "it looks like" and "I'm the queen", where there exists a musical pause.

\subsection{Different Rhyming Difficulties}
\label{sec:app_rhydif}

We noticed that an improper rhyme prompt will lead to lower text quality and a lower chance of constraints being satisfied. For example, Figure~\ref{fig:rhy_dif} shows the rhyme ranking scores of one paragraph and different outputs when using different rhyme targets. With the 1st-ranked rhyme as prompt (Figure~\ref{fig:rhy_dif1}), the output is perfect in length and rhyme control and has a satisfactory translation quality. However, with a rhyme that has a low score (Figure~\ref{fig:rhy_dif2}), the rhyme control performance drops (one missing rhyme), and both sense and naturalness become questionable.

\subsection{Disadvantage of altering beam search}
\label{app:dis_beam}

\begin{figure}[t!]
\centering
\begin{subfigure}{0.48\linewidth}
  \centering
  \includegraphics[width=\textwidth]{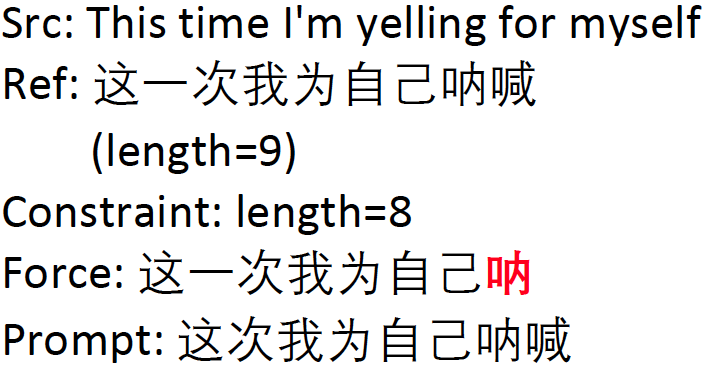}
  \vskip -2mm
  \caption{}
  \label{fig:beam1}
\end{subfigure}
\hfill
\begin{subfigure}{0.48\linewidth}
  \centering
  \includegraphics[width=\textwidth]{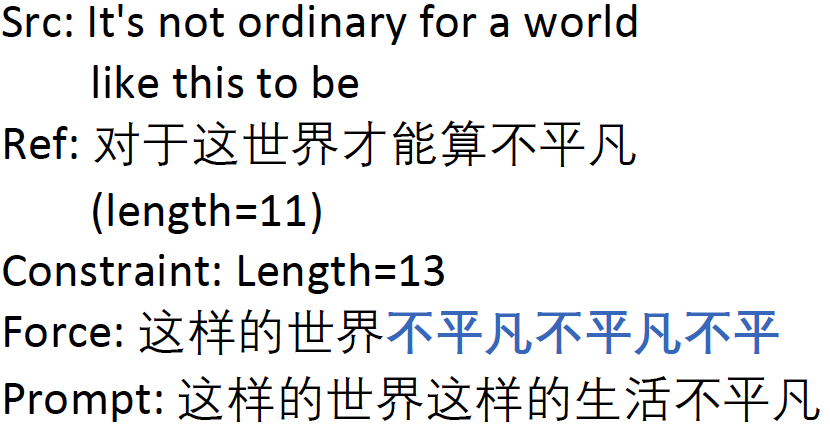}
  \vskip -2mm
  \caption{}
  \label{fig:beam2}
\end{subfigure}

\begin{subfigure}{0.48\linewidth}
  \centering
  \includegraphics[width=\textwidth]{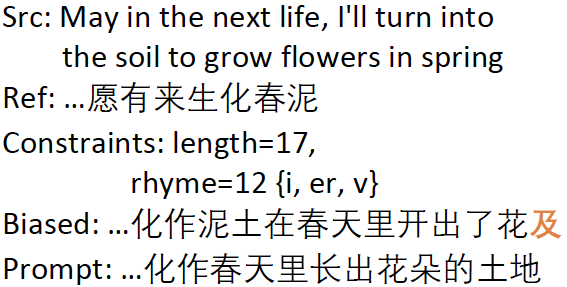}
  \vskip -2mm
  \caption{}
  \label{fig:beam3}
\end{subfigure}%
\hfill
\begin{subfigure}{0.48\linewidth}
  \centering
  \includegraphics[width=\textwidth]{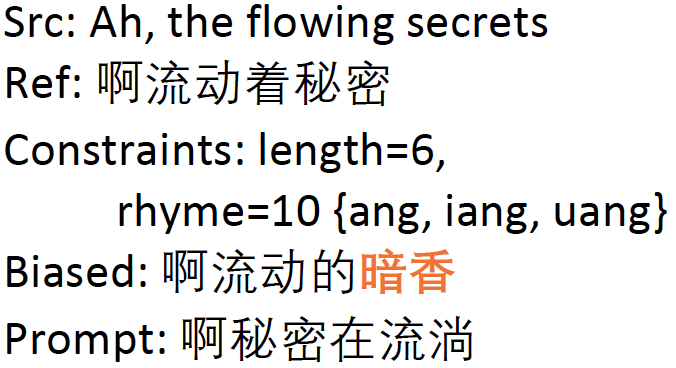}
  \vskip -2mm
  \caption{}
  \label{fig:beam4}
\end{subfigure}

\begin{subfigure}{0.58\linewidth}
  \centering
  \includegraphics[width=\textwidth]{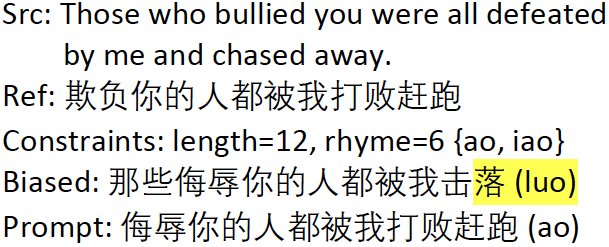}
  \vskip -2mm
  \caption{}
  \label{fig:beam5}
\end{subfigure}
\caption{Comparison of controlling by altering beam search and prompt. (a) and (b): length-controlled translation, where the desired output length is shorter than and longer than reference text length, respectively. (c), (d), and (e): translation with both length and rhyme control, obtained by normal-order and reverse-order decoding, respectively. Text in red: incomplete words; text in blue: repetition; test in orange: words irrelevant to source sentence; highlighted text: wrong rhyme.}
\label{fig:beam}
\vskip -0.1in
\end{figure}

We show the disadvantages of controlling by altering beam search by examples.

\paragraph{Length Forcing} Figures \ref{fig:beam1} and \ref{fig:beam2} show typical errors when the length constraint is different from the length of the reference sentence, which is usually the case at inference time. If the desired length is shorter than the reference, the beam search might end too soon, so the sentence will be incomplete (Figure~\ref{fig:beam1}). If the desired length is longer than the reference (Figure~\ref{fig:beam2}), there tends to be repetition in the outputs. Both cases significantly damage the translation quality, although the outputs may even have higher BLEU scores.

\paragraph{Biased decoding for rhyme} A type of error frequently happens that the end-words in the outputs are biased toward words that satisfy the rhyme constraints but are irrelevant to the source sentences and are incompatible with other parts of the output sentences, as in Figures \ref{fig:beam3} and \ref{fig:beam4}. Such problems are much rarer in translations obtained by prompt-based methods. 

Figure~\ref{fig:beam5} illustrates a possible explanation for the minor performance improvement observed when using a reverse-order decoder with biased decoding for rhyme control. The highlighted word in the biased decoding output, ``落'', has multiple pronunciations. One of these, ``lao'', meets the rhyme requirement. However, the correct pronunciation for this specific context is ``luo'', which does not fulfill the rhyme constraint.

\section{Error Bar}
\begin{figure*}[t!]
\centering
\begin{subfigure}{\linewidth}
  \centering
  \includegraphics[width=\textwidth]{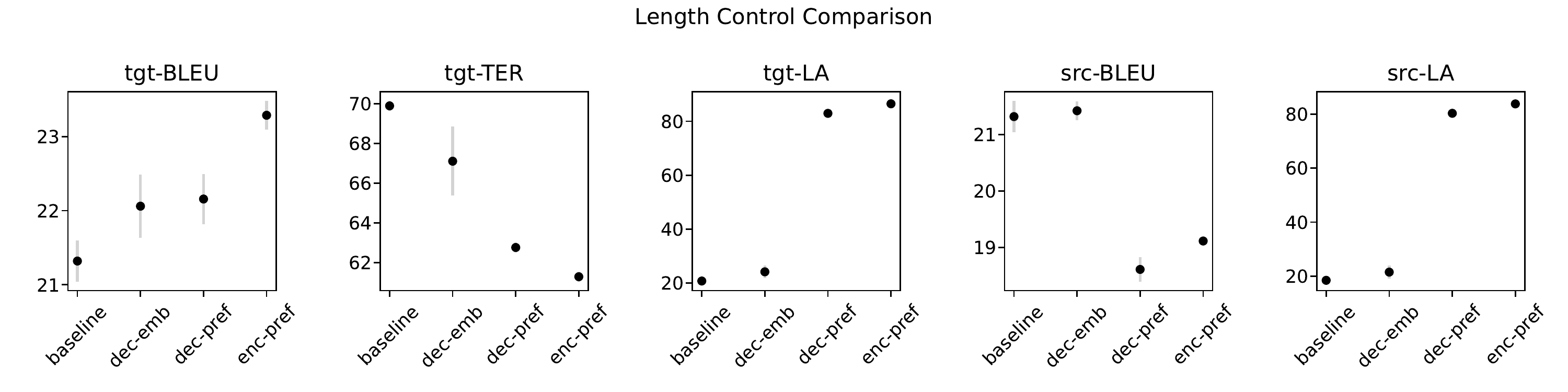}
  \vskip -2mm
\end{subfigure}
\begin{subfigure}{\linewidth}
  \centering
  \includegraphics[width=\textwidth]{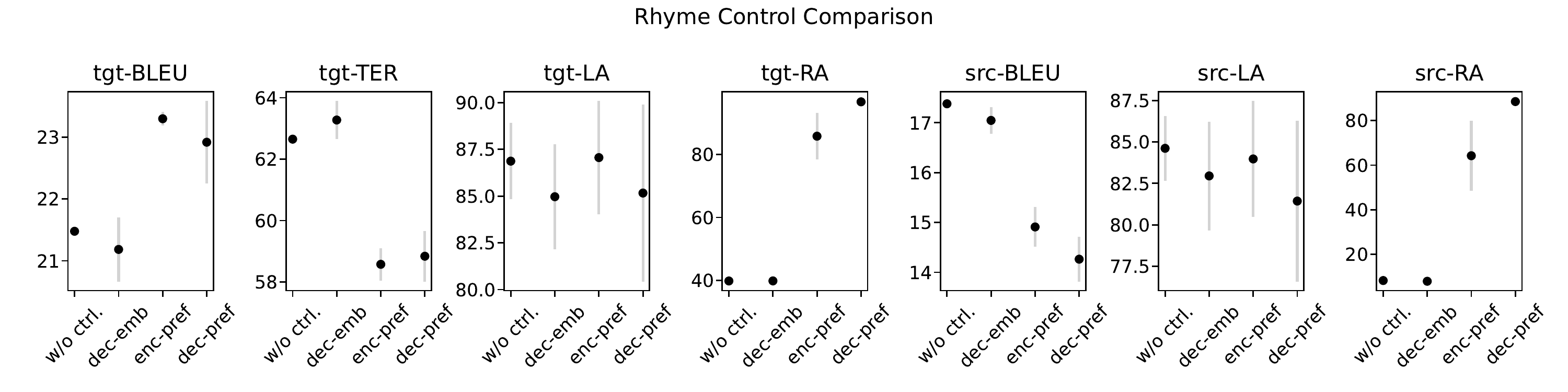}
  \vskip -2mm
\end{subfigure}
\begin{subfigure}{\linewidth}
  \centering
  \includegraphics[width=\textwidth]{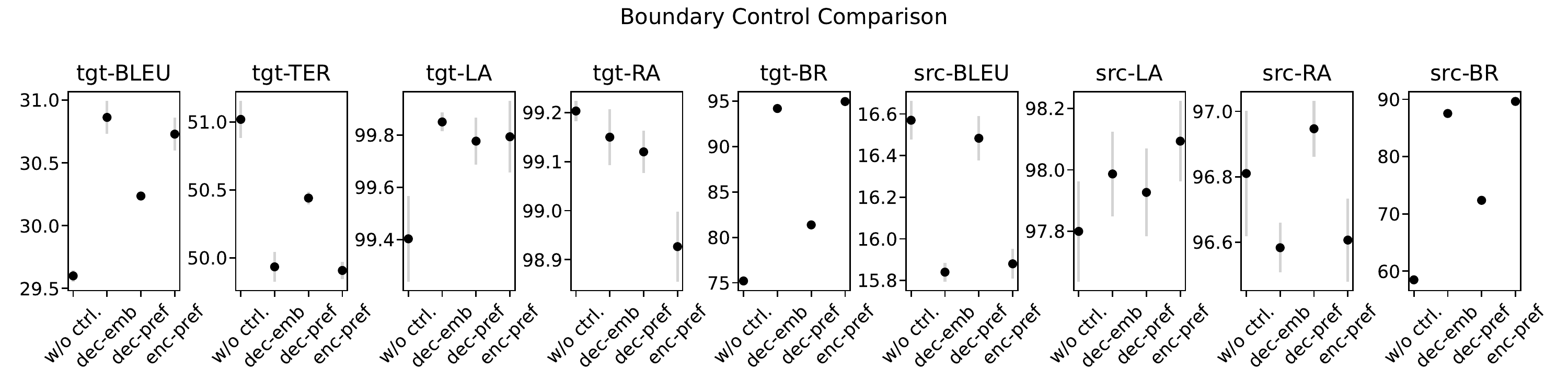}
  \vskip -2mm
\end{subfigure}

\caption{Error bar charts of comparative study of different prompt forms for controlling of different aspects. \textbf{ref-}: results in the tgt-const setting. \textbf{src-}: results in the src-const setting.}

\label{fig:error_bar}
\end{figure*}
\begin{figure*}[!b]
\includegraphics[width=0.8 \linewidth]{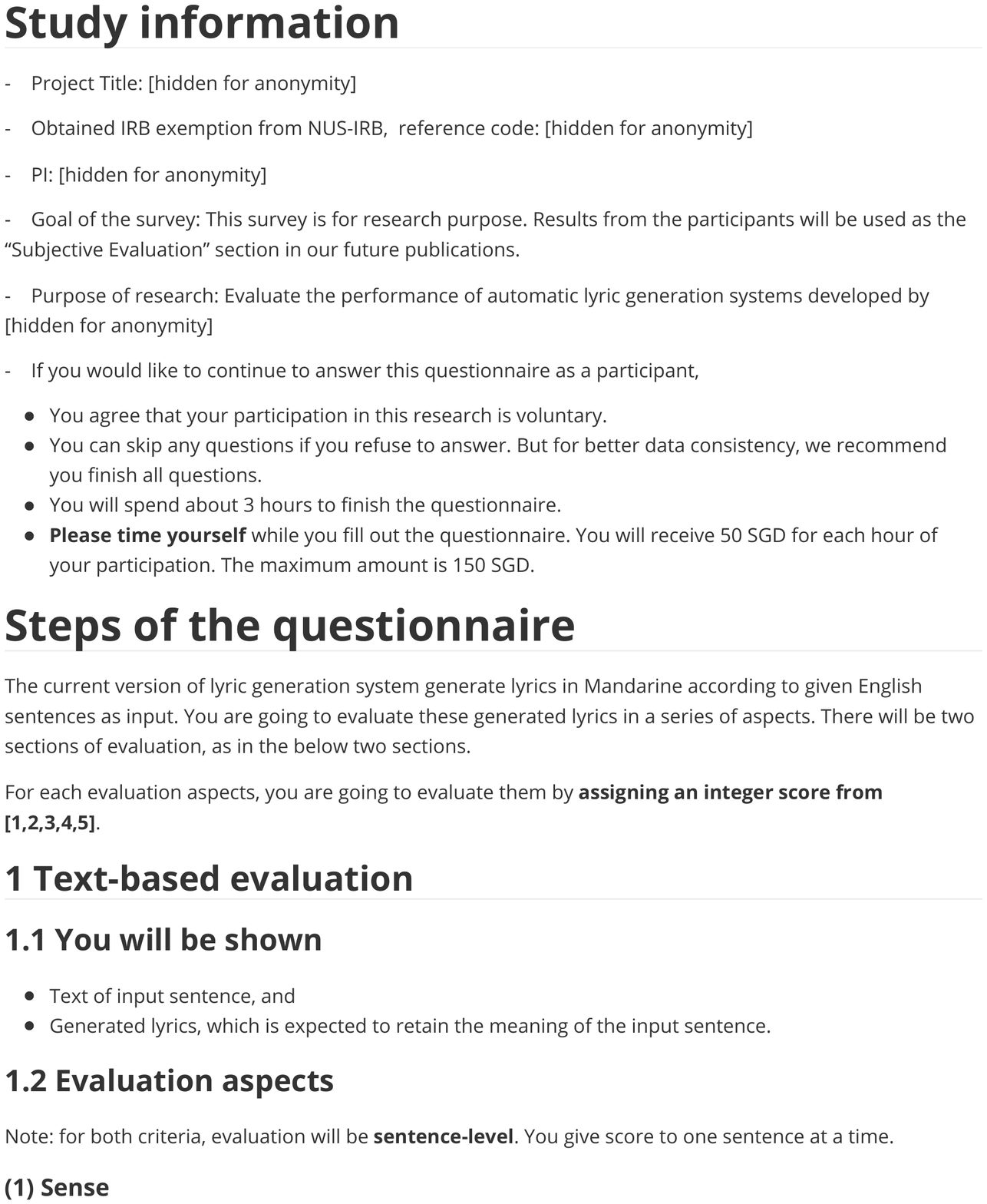}
\centering
\caption{Instructions for human evaluation, page 1/4.}
\label{fig:questionnaire1}
\end{figure*}

\begin{figure*}[!b]
\includegraphics[width=1 \linewidth]{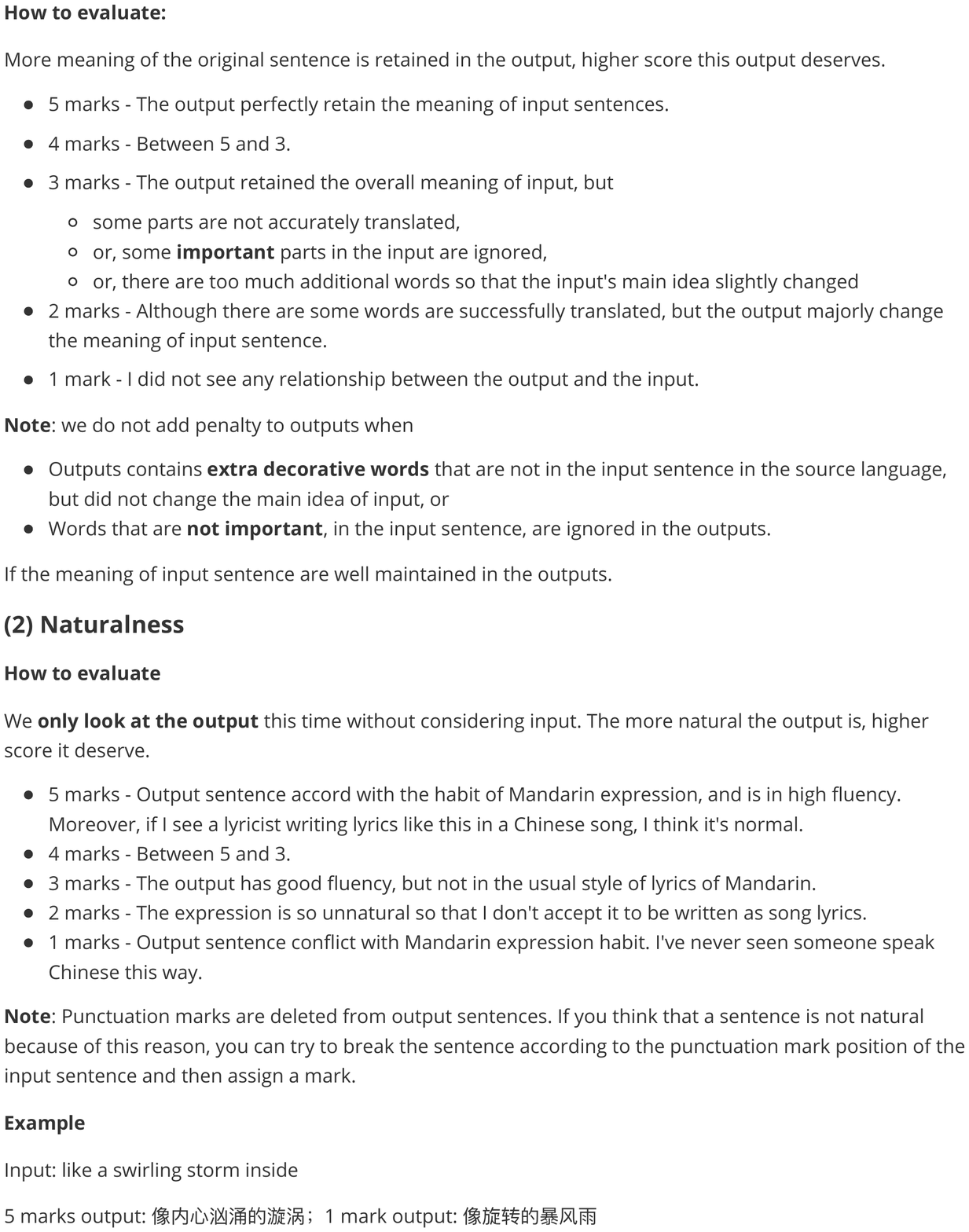}
\centering
\caption{Instructions for human evaluation, page 2/4.}
\label{fig:questionnaire2}
\end{figure*}

\begin{figure*}[!b]
\includegraphics[width=1 \linewidth]{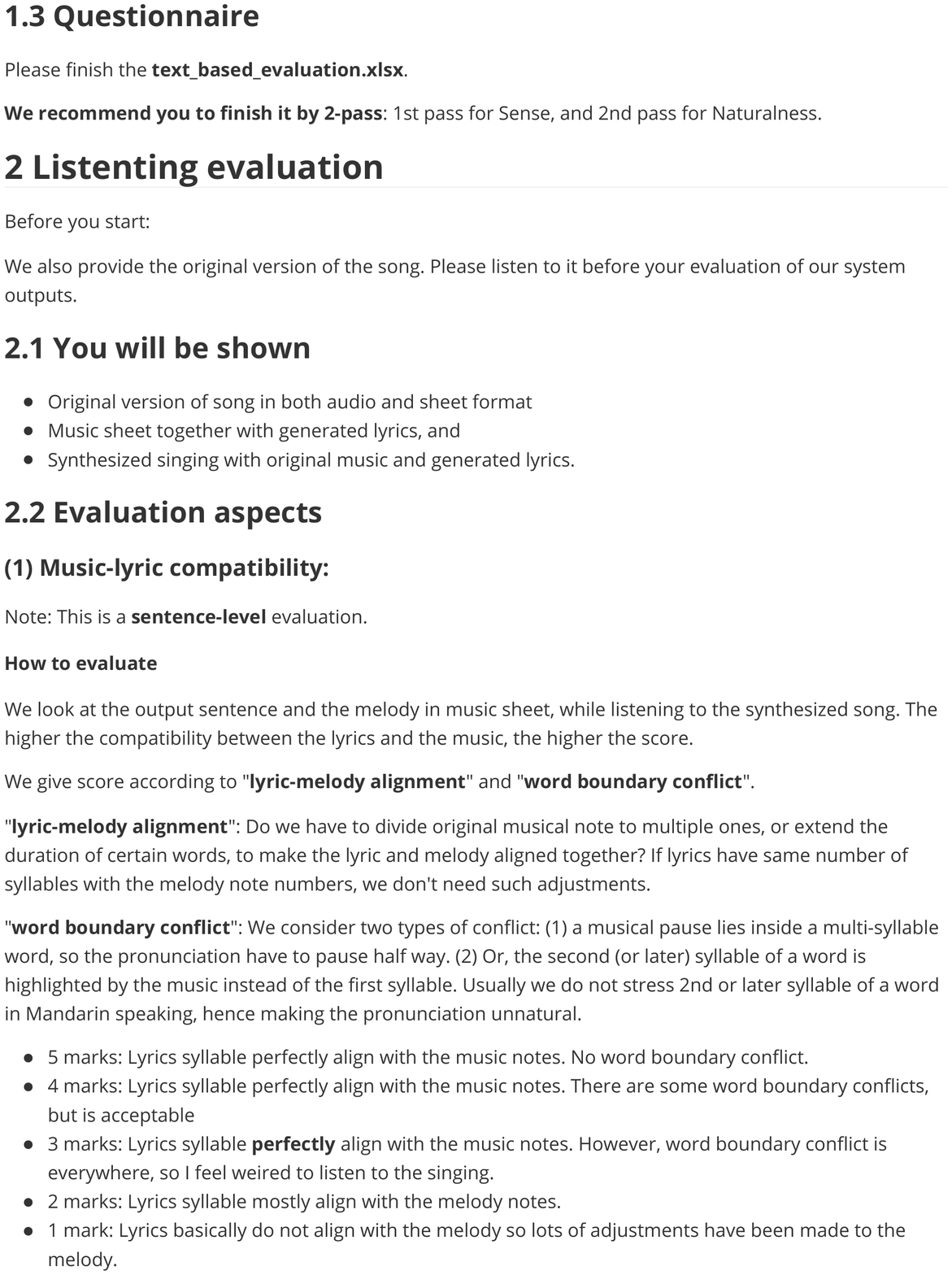}
\centering
\caption{Instructions for human evaluation, page 3/4.}
\label{fig:questionnaire3}
\end{figure*}

\begin{figure*}[!b]
\includegraphics[width=1 \linewidth]{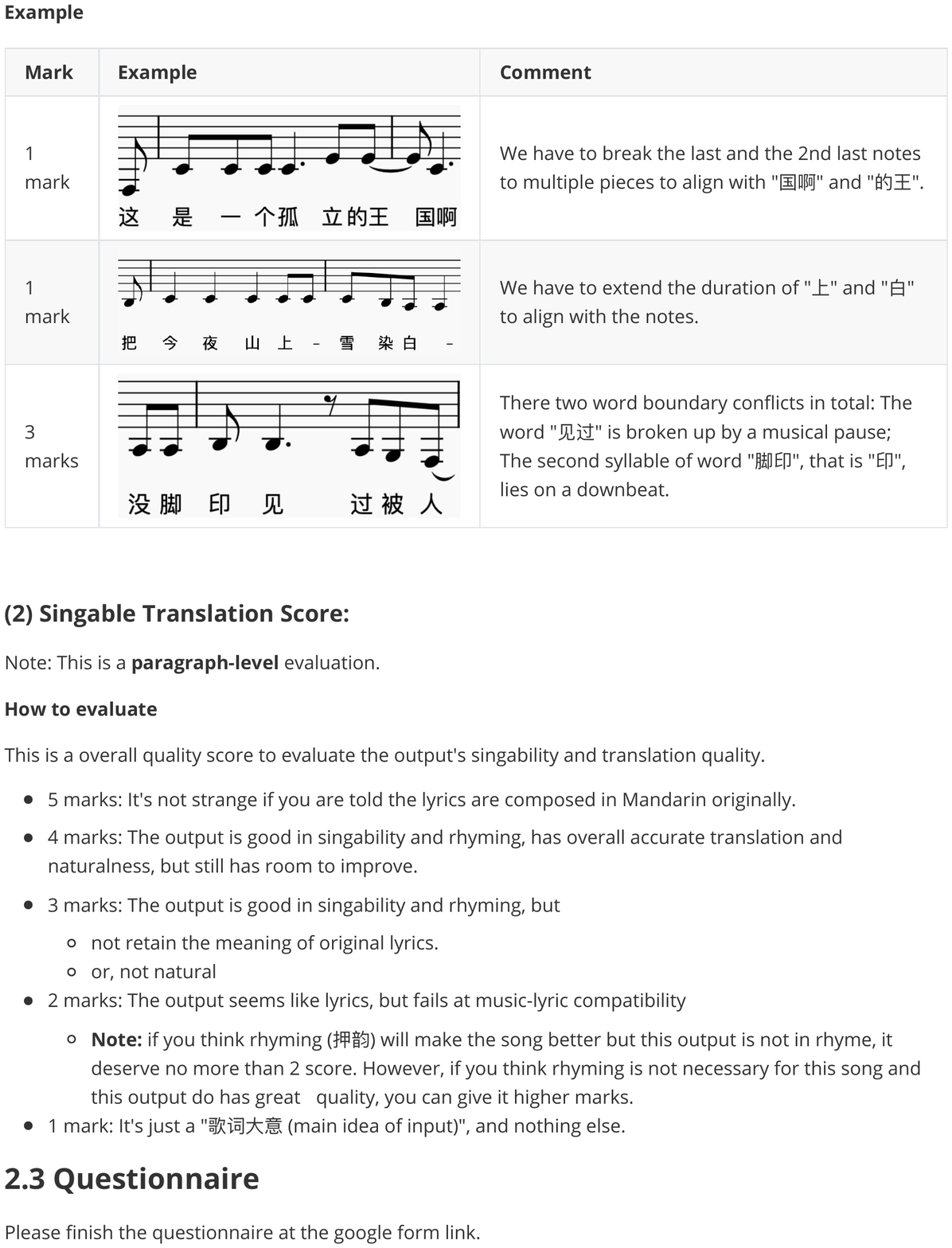}
\centering
\caption{Instructions for human evaluation, page 4/4.}
\label{fig:questionnaire4}
\end{figure*}

In order to reduce the randomness in the results of our comparative study, each experiment in \S\ref{sec:compare} is run three times. Here we show more detailed results by the error bar charts in Figure~\ref{fig:error_bar}.

\section{Subjective Evaluation}
\label{sec:sub_detail}

We select the same five songs as GagaST \cite{guo2022automatic} for our subjective testing. When doing this experiments, we ensure these songs are not in the training set. 

As mentioned in \S\ref{sec:human}, we evaluate the results from four aspects: sense, naturalness, music-lyric compatibility, and the Singable Translation Score (STS), an overall singable translation quality. The four metrics are evaluated at different levels. Sense and naturalness are evaluated for independent text-only sentences, melody compatibility is evaluated for each sentence given the music context, and the last metric is evaluated at the paragraph level. When evaluating STS, we show participants not only the music sheet containing melody notes and lyrics, but also with a singing audio. This audio file contains singing voice synthesized with original melody and generated lyrics, mixed with original musical accompaniments. The voice part is synthesized by ACE Studio\footnote{\url{https://ace-studio.huoyaojing.com/}}. The accompaniments is obtained by using a source separation model Demucs v3 \textit{mdx\_extra} \cite{defossez2019demucs}.

To test the reliability of our subjective metrics, we computed the inter-rater agreement using intra-class coefficients (two-way mixed-effect, average measure model). The results are as follows: 0.8313 for sense, 0.7817 for naturalness, 0.8663 for music-lyric compatibility, and 0.7870 for Singable Translation Score. All of these values fall within the "good reliability" range suggested by \cite{koo2016guideline}. 

\subsection{Instructions For Human Evaluation}

\end{CJK}
\end{document}